\documentclass[letterpaper]{article} 
\usepackage{aaai2026}  
\usepackage{times}  
\usepackage{helvet}  
\usepackage{courier}  
\usepackage[hyphens]{url}  
\usepackage{graphicx} 
\usepackage{natbib}  
\usepackage{caption} 
\usepackage{algorithm}
\usepackage{algorithmic}
\usepackage{booktabs}
\usepackage{subcaption}
\usepackage{multirow}
\usepackage{makecell}
\usepackage{footmisc}
\usepackage{amsmath}
\usepackage{newfloat}
\usepackage{listings}
\DeclareCaptionStyle{ruled}{labelfont=normalfont,labelsep=colon,strut=off} 
\floatstyle{ruled}
\newfloat{listing}{tb}{lst}{}
\floatname{listing}{Listing}

\title{NL2CA: Auto-formalizing  Cognitive Decision-Making from Natural Language Using an Unsupervised CriticNL2LTL Framework}

\author {
    Zihao Deng$^{\rm 1, \rm 2}$,
    Yijia Li$^{\rm 1, \rm 2}$,
    Renrui Zhang$^{\rm 1, \rm 2}$,
    Peijun Ye$^{\rm 1}$\thanks{Corresponding author}
}

\affiliations {
    \textsuperscript{\rm 1}Institute of Automation, Chinese Academy of Sciences\\
    \textsuperscript{\rm 2}School of Artificial Intelligence, University of Chinese Academy of Sciences\\
    \{dengzihao2025, 
    liyijia2023, 
    zhangrenrui2024,
    peijun.ye\}@ia.ac.cn
}

\usepackage{bibentry}
\begin{document}

\maketitle

\begin{abstract}
Cognitive computing models offer a formal and interpretable way to characterize human's deliberation and decision-making, yet their development remains labor-intensive. In this paper, we propose NL2CA, a novel method for auto-formalizing cognitive decision-making rules from natural language descriptions of human experience. Different from most related work that exploits either pure manual or human-guided interactive modeling, our method is fully automated without any human intervention. The approach first translates text into Linear Temporal Logic (LTL) using a fine-tuned large language model (LLM), then refines the logic via an unsupervised Critic Tree, and finally transforms the output into executable production rules compatible with symbolic cognitive frameworks. Based on the resulted rules, a cognitive agent is further constructed and optimized through cognitive reinforcement learning according to the real-world behavioral data. Our method is validated in two domains: (1) NL-to-LTL translation, where our CriticNL2LTL module achieves consistent performance across both expert and large-scale benchmarks without human-in-the-loop feed-backs, and (2) cognitive driving simulation, where agents automatically constructed from human interviews have successfully learned the diverse decision patterns of about 70 trials in different critical scenarios. Experimental results demonstrate that NL2CA enables scalable, interpretable, and human-aligned cognitive modeling from unstructured textual data, offering a novel paradigm to automatically design symbolic cognitive agents.
\end{abstract}



\section{Introduction}
Cognitive architectures such as ACT-R \citep{anderson2014atomic} and Soar \citep{laird2022introduction} are rule-based models, which can simulate various cognitive tasks such as memory, problem solving, and learning. Such architectures have achieved wide applications in human behavior modeling due to their excellent simulation of human cognitive processes and high interpretability. But such architectures require extensive expert effort to instantiate and depend almost exclusively on human experts’ prior knowledge to design. 

Large language models (LLM) such as GPT-4 \citep{achiam2023gpt} and Deepseek \citep{liu2024deepseek}, on the other hand, have shown promising capabilities in imitating human reasoning and behavior after a post-training on corpus with human feedbacks \citep{kumar2025llm}. Even though LLMs are still being criticized for hallucination \citep{rawte2023troubling} and diminishing returns for scaling \citep{villalobos2024position}, their knowledge of human cognitive process can still be used to automate the instantiation of the decision-making model with cognitive architectures \citep{kirk2023exploiting} \citep{niu2024large} \citep{wray2025applying}.


The core component of the cognitive architecture are the declarative memory and procedural memory represented by a set of production rules, each with a precondition and an effect. Cognitive agents operate in perceive–plan–act cycles, dynamically matching environmental features against these rules to determine their actions. There are already some studies on automatically generating production rules using LLM \citep{zhu2024bootstrapping} \citep{kirk2024improving}, but almost all previous work aims to generate the production rules by directly querying the LLM about how the agent should act under an unknown situation. Such approaches construct the cognitive models without human prior knowledge and data, which could lead to a relatively poor alignment with actual human behaviors.

In this work, we propose a novel approach for cognitive agent construction: \textbf{NL2CA} (Natural Language to Cognitive Agent), which leverages LLMs to formalize production rules from natural language descriptions of human experiences. As shown in Figure 1, the LLM is used to interpret textual descriptions of human deliberations and convert them into structured representations. Thus, the LLM serves not merely as a rule-generator, but also as a knowledge-extractor. The generated production rules are based both on the human experience document and the LLM, so they are expected to be more human-aligned.

Realizing this vision requires LLM to have both the capability of providing correct knowledge and the capability of formalizing production rules from human experience documents. Since the former capability is already well demonstrated in previous work, our work focuses mainly on improving the LLM’s capability to formalize production rules from human experience documents.

\begin{figure*}[t]
    \centering
    \includegraphics[width=0.8\textwidth]{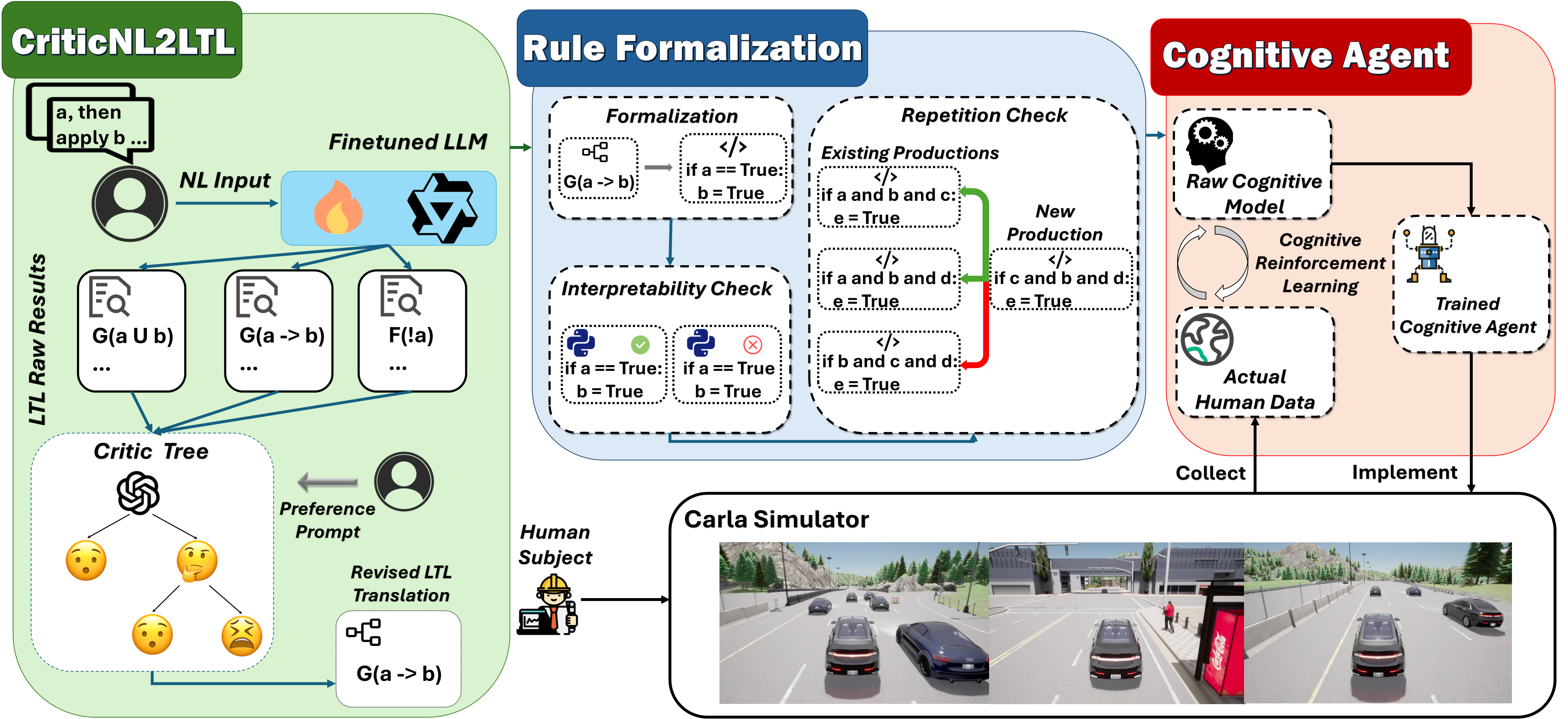}
    \caption{Pipeline of the NL2CA framework.}
    \label{fig:wide}
\end{figure*}

To improve the accuracy and reliability of this production rule extraction, we first prompt the LLM to formalize the logical information of the NL description into Linear Temporal Logic (LTL), which is equivalently a powerful formalism in various domains such as robotics \citep{wang2025conformalnl2ltl}, electronics design \citep{browne1986automatic} and autonomous driving \citep{maierhofer2022formalization}. To facilitate this formalization, We demonstrate a novel framework combining a fine-tuned LLM and a heterogeneous unsupervised Tree-of-Thought (ToT) like Critic Tree structure to improve the NL2LTL translation accuracy. This enhances the generality of our proposed framework, enabling its applicability not only to cognitive modeling, but also to other engineering domains where LTL is operative.

The key contribution of this paper is as follows: 1) we propose a novel framework for fully auto-formalizing cognitive decision-making rules from natural language descriptions of human experience, 2) we propose a more strengthened automated approach to tackle the natural language to linear temporal logic formalization, 3) we examined the validity of our proposed framework in various driving simulation scenarios.

\section{Related Work}
\subsection{Cognitive Architecture Design Automation}
Cognitive architectures (e.g., ACT-R, Soar) have proven to be effective in modeling human cognition but require extensive manual effort. To bridge this burden, recent efforts have tried to leverage LLMs for the instantiation of the cognitive model. STARS \citep{kirk2024improving} and \citet{zhu2024bootstrapping} directly query LLMs to generate production rules for specific environmental states. \citet{bajaj2023generating} proposes a knowledge extraction mechanism, where LLMs are used to populate symbolic memory structures.

While these approaches enable a faster construction of cognitive systems, they often suffer from limited application in actual human cognitive modeling, as those methods are either task-learning based and not grounded in real human decision traces, or simply for knowledge extraction. To the best of our knowledge, there is still no work on automating the human cognitive modeling pipeline. Our work addresses these limitations by formalizing natural language descriptions of human experience into LTL, preserving both traceability and human-likeness, and transforming it into executable production rules for cognitive modeling.

\subsection{Cognitive Driving Model}
The idea of combining cognitive architecture to imitate driver behaviors was first brought up by \citet{salvucci2006modeling}. They proposed a general pipeline of how to model driver behaviors by using a specific cognitive architecture and introduced a threaded cognition for driving tasks.

Later, some approaches were conducted to further extend cognitive architectures’ application to the construction of virtual drivers. \citet{ritter2019act} elucidated the operation of ACT-R, describing how driver deals with information and generates behaviors through monitoring, decision-making and control processes. \citet{oh2024driver} proposed an approach to predict mental workload during take-over situations in autonomous driving. Although numerous attempts were made, due to their lack of learning capability, early approaches were often limited to singular scenarios, thus their applications to autonomous driving are still limited \citep{chater2018negotiating}.

To address these issues, \citet{qi2024cognitive} proposed a cognitive reinforcement learning method to align the cognitive reasoning via human-machine interactions. \citet{li2024interpretable} further refined such method and reported an interpretable cognitive driving model. Their method was reported to outperform standard PPO algorithm, while offering a highly interpretable pattern. 

However, these methods require manual specification of production rules, limiting their extensibility. In contrast, our NL2CA framework automates the pipeline by directly formalizing production rules from NL data, making it more extensible to multiple scenarios without manual inspections.

\section{Method}
The NL2CA framework consists of three stages: 1) Formalizing logic representations from NL text using a fine-tuned LLM and an unsupervised Critic Tree, 2) Transforming the resulted temporal logic into executable cognitive production rules, and 3) Training a cognitive agent through cognitive reinforcement learning (CRL) on human data.

In the first stage, an input NL segment is formalized as a Linear Temporal Logic (LTL) formula. The Critic Tree module iteratively refines the formula to better align with the meaning of the original sentence. The refined LTL is then grounded with variables from a predefined knowledge base and transformed into production rules compatible with cognitive modeling frameworks.

In the final stage, a cognitive symbolic-driven agent is constructed based on the generated production rules. The agent is trained on real human behavioral data using cognitive reinforcement learning method, adjusting the rules’ utilities based on how closely its outputs match human behaviors. Figure 1 illustrates the NL2CA pipeline.

\subsection{CriticNL2LTL Framework}
To enable an accurate and unsupervised  NL to LTL formalization, we introduce an CriticNL2LTL module, which consists of two main components: a fine-tuned LLM for initial translation and a heterogeneous Critic Tree for iterative refinement, which is shown in Algorithm 1.

Given an NL input, the fine-tuned LLM first generates a raw LTL formula. This formula is then passed to the Critic Tree, a self-organizing revision structure with user-configurable parameters like maximum depth and number of critics. The Critic Tree consists of a revisor LLM, which is responsible for revising the current formula, and critic LLMs, which evaluate the formula's logical correctness and its alignment with user-defined preferences (e.g., grounded knowledge base, desired format). When a critic detects a mismatch, it appends a feedback message to the context of the revisor, instructing the revisor to revise the formula and split a new child node, the child node stores the revised LTL formula and the context of the revisor. The process repeats until all critics approve and no new child nodes can be created or the maximum depth is reached. We construct the critics using heterogeneous LLMs to provide diverse suggestions for revision, since standard Self-refine approaches are reported to fail to improve the accuracy of NL-to-LTL translation \citep{fang2025enhancing}.

\begin{algorithm}[tb]
\caption{Critic Tree}
\textbf{Input}: Natural language text $T$, Initial translation $A_{\text{init}}$ \\
\textbf{Parameters}: Number of critics $\delta$, Maximum depth $D$ \\
\textbf{Output}: Final translation $A_{\text{final}}$
\begin{algorithmic}[1]

\STATE $(A_{\text{root}}, C_{\text{root}}) \leftarrow \text{Revisor}(T, A_{\text{init}})$
\STATE Initialize root node $v_0 = (A_{\text{root}}, C_{\text{root}})$
\STATE Initialize node list $\mathcal{N} \leftarrow \{v_0\}$

\FOR{$d = 0$ to $D$}
    \STATE $\mathcal{N}_{\text{new}} \leftarrow \emptyset$
    
    \FOR{\textbf{each} node $v=(A, C)$ in $\mathcal{N}$}
        \IF{$v$ already has children}
            \STATE \textbf{continue}  
        \ENDIF
        \STATE $results \leftarrow \emptyset$
        \FOR{$k = 1$ to $\delta$}
            \STATE $r_k \leftarrow \text{Critic}(A, T)$
            \STATE $results.\text{append}(r_k)$

            \IF{\textbf{all} $results$ indicate approval}
                \STATE $A_{\text{final}} \leftarrow A$
                \STATE \textbf{return} $A_{\text{final}}$
            \ENDIF
            \STATE $C' \leftarrow C \cup r_k$
            \STATE $(A', C'') \leftarrow \text{Revisor.revise}(C')$
            \STATE Add new node $v' = (A', C'')$ to $\mathcal{N}_{\text{new}}$
        \ENDFOR
    \ENDFOR
    
    \STATE $\mathcal{N} \leftarrow \mathcal{N}_{\text{new}}$
\ENDFOR
\STATE $A_{\text{final}} \leftarrow A_{\text{root}}$
\STATE \textbf{return} $A_{\text{final}}$

\end{algorithmic}
\end{algorithm}

\subsection{Logic to Code Transformation}
After the logic formalization, the LTL formulas are transformed into executable production rules compatible with \textbf{pyactr}, an open source Python library used to create and run cognitive models. We leverage LLM to align the propositions of the LTL formulas with computational variables and to ensure the formalization. This step also includes a post-process to ensure the logical correctness, interpretability, and non-redundancy.

To address common issues such as Function interface mismatch, Over-constraining/generalization and Rule duplication, reported by \citet{zhu2024bootstrapping}, we implement the following mechanisms:

\begin{itemize}
\item \textbf{Function interface mismatch}: The generated code is immediately interpreted using pyactr. If execution fails, the error message and the faulty rule are returned to the LLM for revision via an error-guided correction loop.
\item \textbf{Over-constraining / over-generalization}: Most logical inconsistencies are filtered by the CriticNL2LTL module. However, ambiguous or contradictory input texts may still yield problematic rules. These are handled during the cognitive reinforcement learning stage when rules are evaluated against actual human behaviors.
\item \textbf{Rule duplication}: To avoid redundant rules, each rule is assigned a name based on its preconditions and effects. We transform the rule names into BERT embeddings and use cosine similarity to measure affinity. Each newly generated rule will be compared with existing rules with top 5 similarity scores. If a newly generated rule matches an existing one semantically, it is discarded.
\end{itemize}

These measures ensure that only interpretable, distinct, and executable production rules are retained for cognitive agent modeling.

\begin{table*}[t]
\centering
\begin{tabular}{llcccc}
\toprule
\textbf{Strategy} & \textbf{Model} & \multicolumn{2}{c}{\textbf{$Dataset_1$}} & \multicolumn{2}{c}{\textbf{$Dataset_2$}} \\
 & & \textbf{$Acc^*$ (\%)} & \textbf{$BLEU$} & \textbf{$Acc$ (\%)} & \textbf{$BLEU$} \\
\midrule
\textit{Interactive} & GPT3.5 & 75.0 & 0.814 & 63.3 & 0.916 \\
 & GPT4 & \textbf{94.4} & \textbf{0.911} & 78.3 & \textbf{0.978} \\
 & Gemini & 88.9 & 0.884 & 70.4 & 0.960 \\
\midrule
\textit{Fine-tuning} & Flan-t5 & 8.3 & 0.181 & 90.8 & 0.987 \\
 & Qwen3-0.6B & 22.2 & 0.246 & \textbf{97.5} & \textbf{0.993} \\
\midrule
\textit{End-to-End} & Deepseek R1 & 63.9 & 0.546 & -- & -- \\
 & +Self-Refine & 63.9 & 0.490 & -- & -- \\
\midrule
\textit{Unsupervised} & \textbf{CriticNL2LTL (ours)} & \underline{80.6} & \underline{0.517} & \underline{90.2} & \underline{0.945} \\
\bottomrule
\end{tabular}
\caption{Comparative Performance Analysis}
\end{table*}

\subsection{Cognitive Reinforcement Learning}
The production rules extracted from natural language may still contain inconsistencies or errors due to ambiguities in the input. To improve reliability, we apply cognitive reinforcement learning to adjust each rule's utility based on its alignment with human decision-making behaviors.

As mentioned above, cognitive decision-making is about dynamically matching relevant environmental features to the production rules and applying their effects. Some states may fire multiple production rules, thus leading to many reasoning branches. To diminish the ambiguity, the activation probability for each production rule $p_i$ at step $t$ will be determined by calculating its normalized utility values $u_i$ :
\begin{equation}
p(q_t = p_i) = \frac{e^{u_i / \sigma}}{\sum_j e^{u_j / \sigma}}
\end{equation}
where $\sigma$ is a super parameter.
During the cognitive agent training process, we are mainly concerned about whether its final decision aligns with the actual human decision in the same situations. So, each time the model outputs a same result as actual human decision, its reasoning chain receives a reward $R_+$, otherwise a negative reward $R_-$ is assigned, the reward will be decomposed into all activated production rules on the previous reasoning path:
\begin{equation}
r_i = R - \beta \Delta t
\end{equation}
where $\Delta t$ represents the time gap between the activation of production rule $p_i$ and the moment the reward is received, while $\beta$ denotes the decay coefficient, reflecting that production rules triggered earlier receive smaller portions of the reward. Subsequently, each production rule's utility is updated based on its allocated reward: 
\begin{equation}
u_i(n) = u_i(n-1) + \alpha \left[ r_i(n) - u_i(n-1) \right]
\end{equation}
where $n$ denotes the number of iterations, and $\alpha$ denotes the learning rate. If the reward a production rule receives exceeds its current utility, the utility will be increased, thereby enhancing its probability of being activated under similar conditions. Conversely, if the reward is lower, the utility will be reduced.

\subsection{LTL and Production Rules}
In this section, we will clarify how we ensure the formalization of LTL to production rules.

LTL is built upon classical propositional logic, but with temporal operators introduced, it can also describe sequences of events and temporal properties. Unlike CTL (Computation Tree Logic), LTL only focuses on linear time evolutions along individual execution paths. The syntax of LTL is as follows:
\begin{equation}
\varphi ::= true \mid a \mid \varphi_1 \land \varphi_2 \mid \neg \varphi \mid \bigcirc \varphi \mid \varphi_1 \cup \varphi_2
\end{equation}
where $true$ refers to a logical constant that is always satisfied, $a$ refers to an atomic proposition representing a basic statement about the system, $\land$ stands for conjunction operator, $\neg$ stands for negation operator, $\bigcirc$ is a temporal operator which means that the proposition must be true in the next state or moment in time, $\varphi_1 \cup \varphi_2$ expresses that $\varphi_1$ must remain true continuously until $\varphi_2$ becomes true at some point in the future.

Production rules, on the other hand, are commonly described in the “$IF <condition> THEN <action>$” format, expressing a simple conditional logic: whenever the condition is met, the associated action is triggered. With that, the production rules can be seen as a subset of LTL. An LTL formula of “$G$( a $\rightarrow$ b $)$”, describing the logic of “whenever a holds, b holds as well”, can be formalized as “$IF$ a $THEN$ b”, describing the same logic.

But not all LTL formulas can be formalized as production rules. The operators like “$F$” (finally), “$\cup$” (until) can’t find their synonyms in production rules. The meaning of such temporal operators go beyond what production rules can represent. If a segment of NL text is formalized into LTL formula containing these temporal operators, it implies that the text involves overly complex logic and is therefore not suitable for further formalization into production rules. So, in our work, the LTL formulas using such operators will be marked as \textbf{Inference Error}, and only LTL formulas in the format of  “$G(<propositions>$ $\rightarrow$ $<propositions>$” are formalized into production rules.

Introducing LTL as an intermediate representation offers several benefits: 1) It transforms the unstructured natural language descriptions of human experience into well-defined logical formula, making it easier to extract and formalize corresponding production rules. 2) It allows for explicit representation of complex logic that can’t be expressed as production rules, preventing inappropriate results. 3) LTL abstracts away from domain-specific actions, it can serve as a reusable, task-agnostic layer that facilitates generalization across different tasks or scenarios.

\section{Experiment}
To validate the proposed NL2CA framework, we need to examine its capabilities in two aspects:

\begin{itemize}
\item Can CriticNL2LTL framework extract accurate logic information from natural language texts? (see section \textbf{NL-to-LTL Experiment})
\item Can NL2CA construct a cognitive agent which is capable of imitating human behaviors? (see section \textbf{Cognitive Driving Simulation})
\end{itemize}

\subsection{NL-to-LTL Experiment}
We conduct NL-to-LTL experiments on two datasets: $Dataset_1$, a challenging open-source benchmark for neural translation approach \citep{cosler2023nl2spec}, and $Dataset_2$, a larger-scale public synthetic dataset \citep{xu2024learning}, containing samples across eight LTL expression patterns and various domains.

We fine-tune a Qwen3-0.6B model on the training samples on $Dataset_2$ to generate initial LTL translations, which are then refined using our Critic Tree constructed with Deepseek models (number of critics = 2, max depth = 2),  the revisor is constructed using DeepseekR1. For critics, each critc is constructed with a 50\% probability based on DeepseekR1 and a 50\% probability based on DeepseekV3. Performance is assessed using the match accuracy ($ACC$) and $BLEU$ score. For $Dataset_1$, due to its smaller scale and high level of difficulty, accuracy is evaluated through manual semantic verification (denoted as $ACC^*$). To address generality, we only use the $minimal$ prompt that only includes minimal domain knowledge (details in Appendix). In practice, it can be replaced with more domain-specific content to further boost the framework’s performance.


The experimental results are shown in Table 1. For LLM-based methods, the performance on $Dataset_2$ is measured on 143 samples randomly picked from the test dataset.

Our approach achieves high consistency across both datasets, outperforming purely fine-tuned and end-to-end approaches on $Dataset_1$, while maintaining competitive results on $Dataset_2$. Interactive approaches that benefit from human oversight perform outstandingly on $Dataset_1$ but show a significant performance drop on $Dataset_2$ due to domain shift. Notably, conventional self-Refine \citep{madaan2023self} fails to improve the performance of end-to-end baselines, highlighting the effectiveness of our proposed Critic Tree structure.

\begin{figure}[t]
\centering

\begin{subfigure}{0.45\linewidth}
  \centering
  \includegraphics[width=\linewidth]{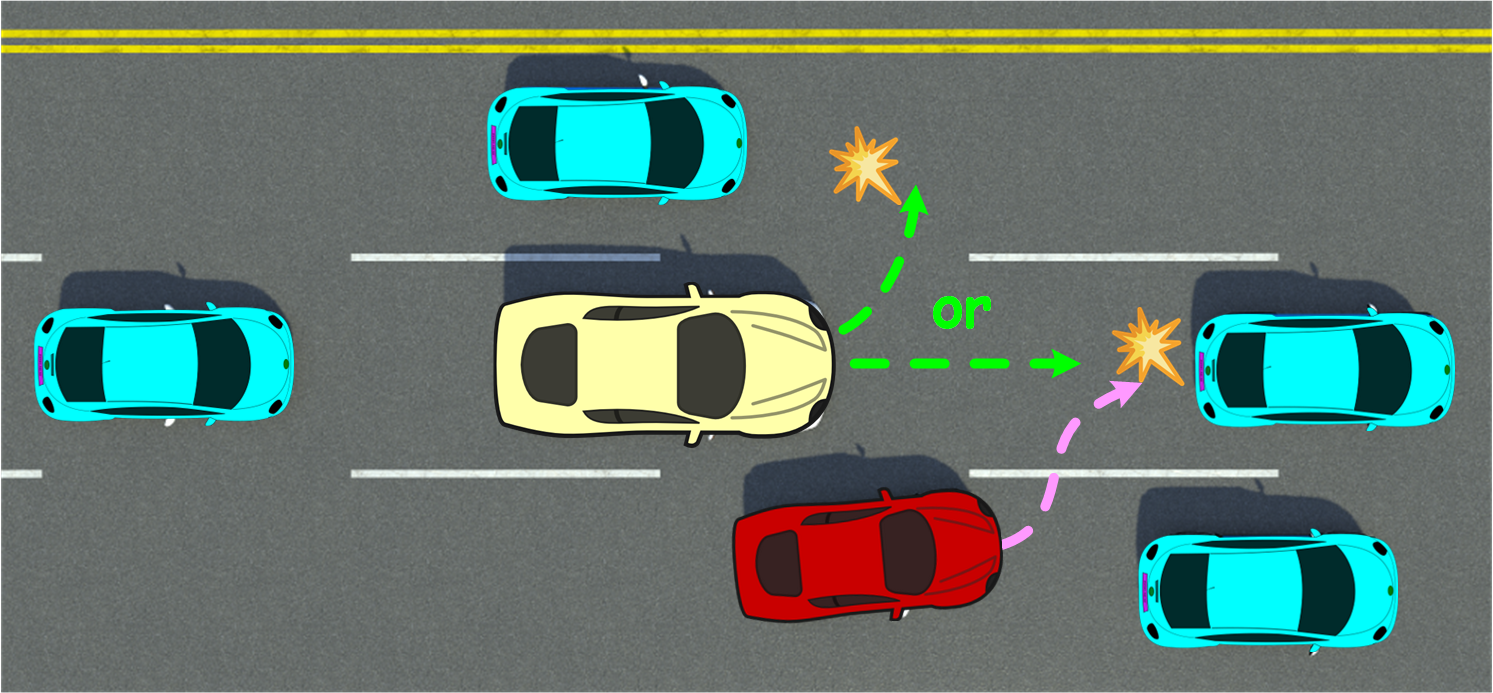}
  \caption{Highway}
  \label{fig:img1}
\end{subfigure}
\hspace{0.05\linewidth}
\begin{subfigure}{0.45\linewidth}
  \centering
  \includegraphics[width=\linewidth]{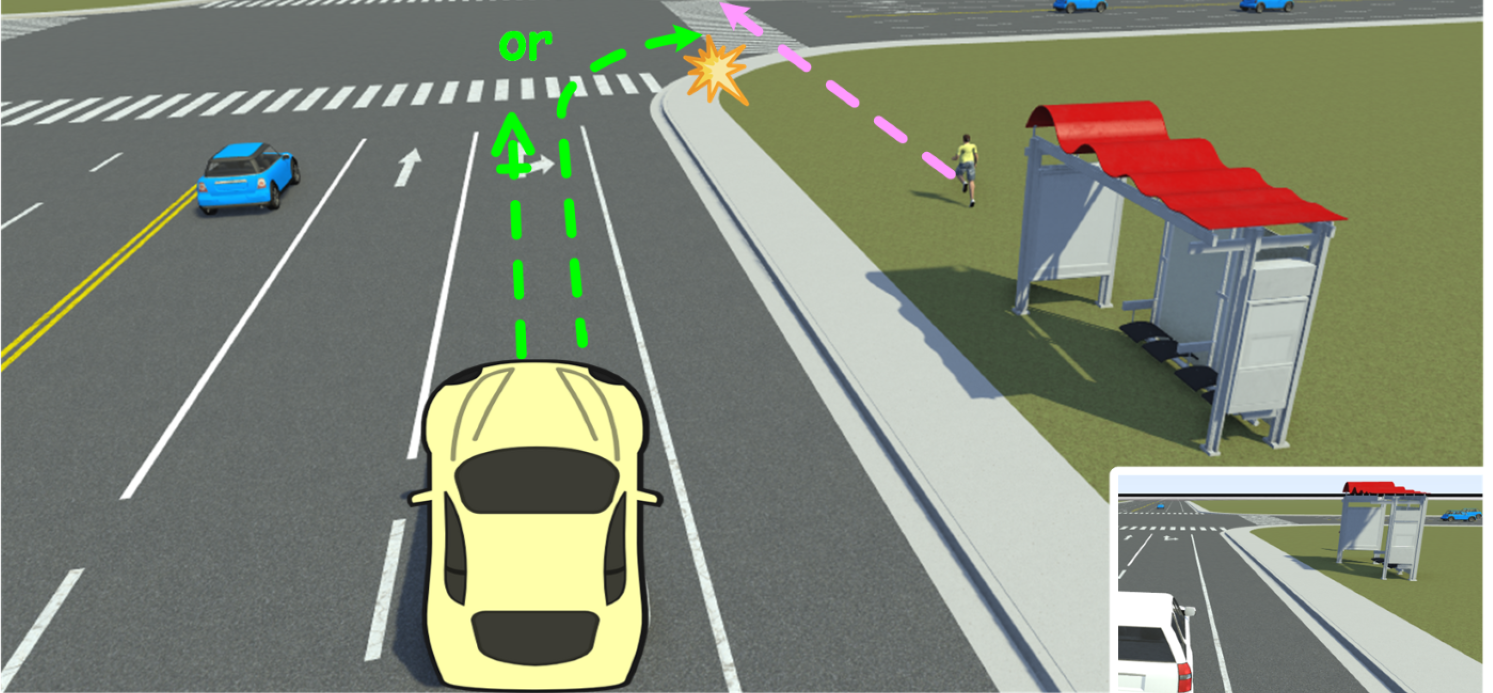}
  \caption{Signalized Intersection}
  \label{fig:img2}
\end{subfigure}

\begin{subfigure}{0.45\linewidth}
  \includegraphics[width=\linewidth]{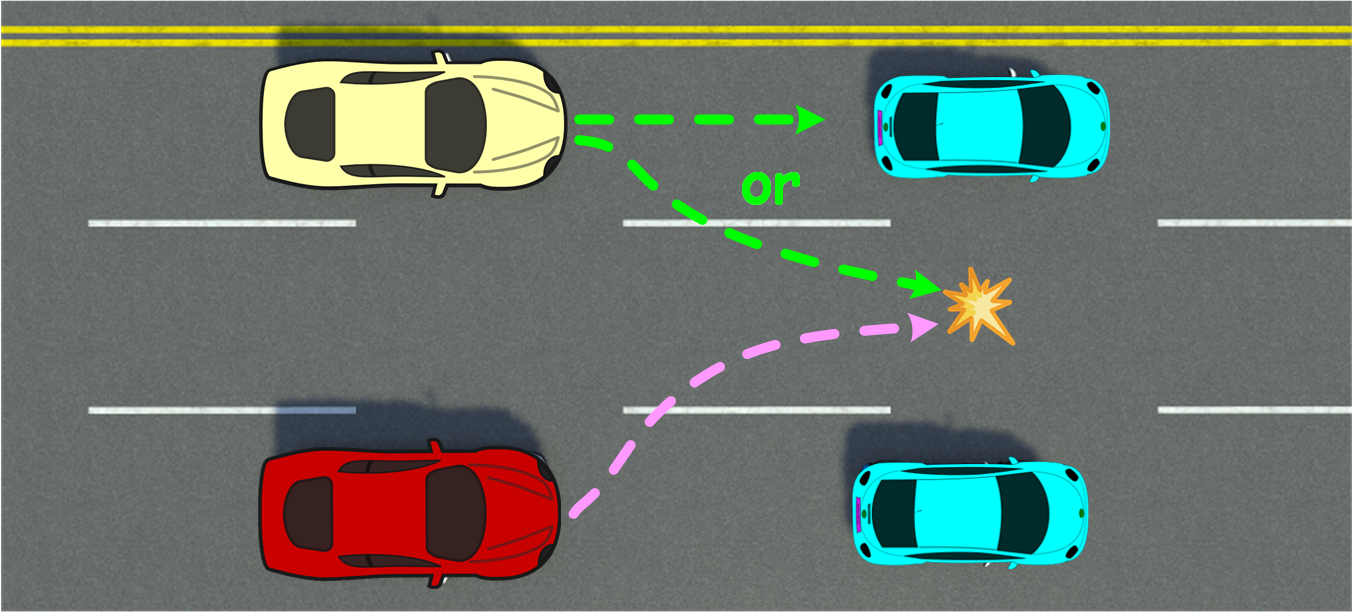}
  \caption{Lane Changing Interference}
  \label{fig:img3}
\end{subfigure}

\caption{Examples of Driving Simulation Scenarios.}
\label{fig:triple}
\end{figure}

\subsection{Cognitive Driving Simulation}
To verify NL2CA’s capability of constructing proper cognitive models from human experience texts, we conduct several cognitive driving simulation experiments. We construct 3 representative driving scenarios including highway, signalized intersection and lane changing interference using CARLAUE4, as shown in Figure 2: 

\begin{itemize}
\item \textbf{Highway}: A three-lane highway scenario, there is a group of vehicles ahead of the ego vehicle occupying all three lanes. During the driving process, an NPC vehicle to the right of the ego-vehicle will attempt to cut in to the ego-vehicle’s lane.
\item \textbf{Signalized Intersection}:A signalized intersection scenario, the ego-vehicle needs to make a right turn through the intersection. During the process, a pedestrian NPC will attempt to cross the road, hindering the ego-vehicle.
\item \textbf{Lane Changing Interference}: A three-lane highway scenario, the ego-vehicle is driving in the leftmost lane, with a slow-driving NPC vehicle ahead. If the ego-vehicle attempts to overtake, an NPC vehicle in the rightmost lane will attempt to make a left lane change.
\end{itemize}

\begin{table}[t]
\centering
\begin{tabular}{ccc}
\toprule
\textbf{Parameter} & \textbf{Description} & \textbf{Value} \\
\midrule
$\alpha$ & Learning rate & 2e-4 \\
$\beta$ & Attenuation coefficient & $0.01$ \\
$\sigma$ & Utility coefficient & $\sqrt{2}$\\
$u_0$ & Initial utility of production& $0$ \\
$R_+$ & Positive reward & $10$ \\
$R_-$ & Negative reward & $0$ \\
\bottomrule
\end{tabular}
\caption{Experimental Parameters}
\end{table}

These scenarios are carefully designed to represent corner cases in urban and highway driving, where decision-making is highly sensitive to subtle environmental changes and driver intent. By focusing on these cases, we aim to assess whether the cognitive agents constructed through NL2CA can capture a general human reasoning patterns.

\begin{table*}[t]
\centering
\label{tab:driving-sim}
\begin{tabular}{llccccc}
\toprule
\textbf{Scenario} & \textbf{Model} & \textbf{Success Rate (\%)$\uparrow$} & \textbf{RSR (\%)$\uparrow$} & \textbf{ACR (\%)$\uparrow$} & \textbf{AST (s)} & \textbf{ALD (cm)$\downarrow$} \\
\midrule
\multirow{4}{*}{\makecell[l]{Highway}}  
 & Human Data & 53.73 & -- & 83.98 & 27.79 & 27.86 \\
 & Manual & 25.00 & \textbf{98.67} & \textbf{89.06} & 26.29 & 25.78 \\
 & $M_{lit}$ & \textbf{58.70} & 87.62 & 82.02 & 26.41 & \textbf{9.43} \\
 & $M_{sup}$ & 12.00 & 93.07 & 77.58 & 22.29 & 10.70 \\
\midrule
\multirow{4}{*}{\makecell[l]{Signalized\\Intersection}}  
 & Human Data & 84.13 & -- & 99.56 & 16.08 & 31.93 \\
 & Manual & 75.00 & \textbf{100.00} & 92.41 & 10.19 & 22.99 \\
 & $M_{lit}$ & \textbf{85.71} & 99.46 & \textbf{99.84} & 7.25 & \textbf{20.15} \\
 & $M_{sup}$ & 76.60 & 99.62 & 99.77 & 8.83 & 22.26 \\
\midrule
\multirow{4}{*}{\makecell[l]{Lane Changing\\Interference}}  
 & Human Data & 85.92 & -- & 97.59 & 28.71 & 39.88 \\
 & Manual & \textbf{95.92} & 99.73 & \textbf{97.81} & 22.42 & \textbf{25.54} \\
 & $M_{lit}$ & 91.30 & \textbf{100} & 89.73 & 30.87 & 35.05 \\
 & $M_{sup}$ & 63.82 & \textbf{100} & 89.70 & 19.19 & 21.65 \\
\bottomrule
\end{tabular}
\caption{Comparative Driving Simulation Analysis}
\end{table*}

We also invite 34 different drivers and conduct about 70 trials in total for each virtual scenario using Virtual-Reality device. They are asked to drive from the predefined start point to the destination, along with complying with the traffic rules and avoiding any potential collisions. During the driving process, their control input (e.g., throttle, steering angle, etc.) and the environmental state information are recorded as the actual human data used for cognitive reinforcement learning. After the driving process, we interview each driver about the driving decisions they made and the corresponding reasons, thus collecting a set of human driving experience texts for logic information extraction. 

\begin{figure}[t]
    \centering
    \includegraphics[width=0.5\textwidth]{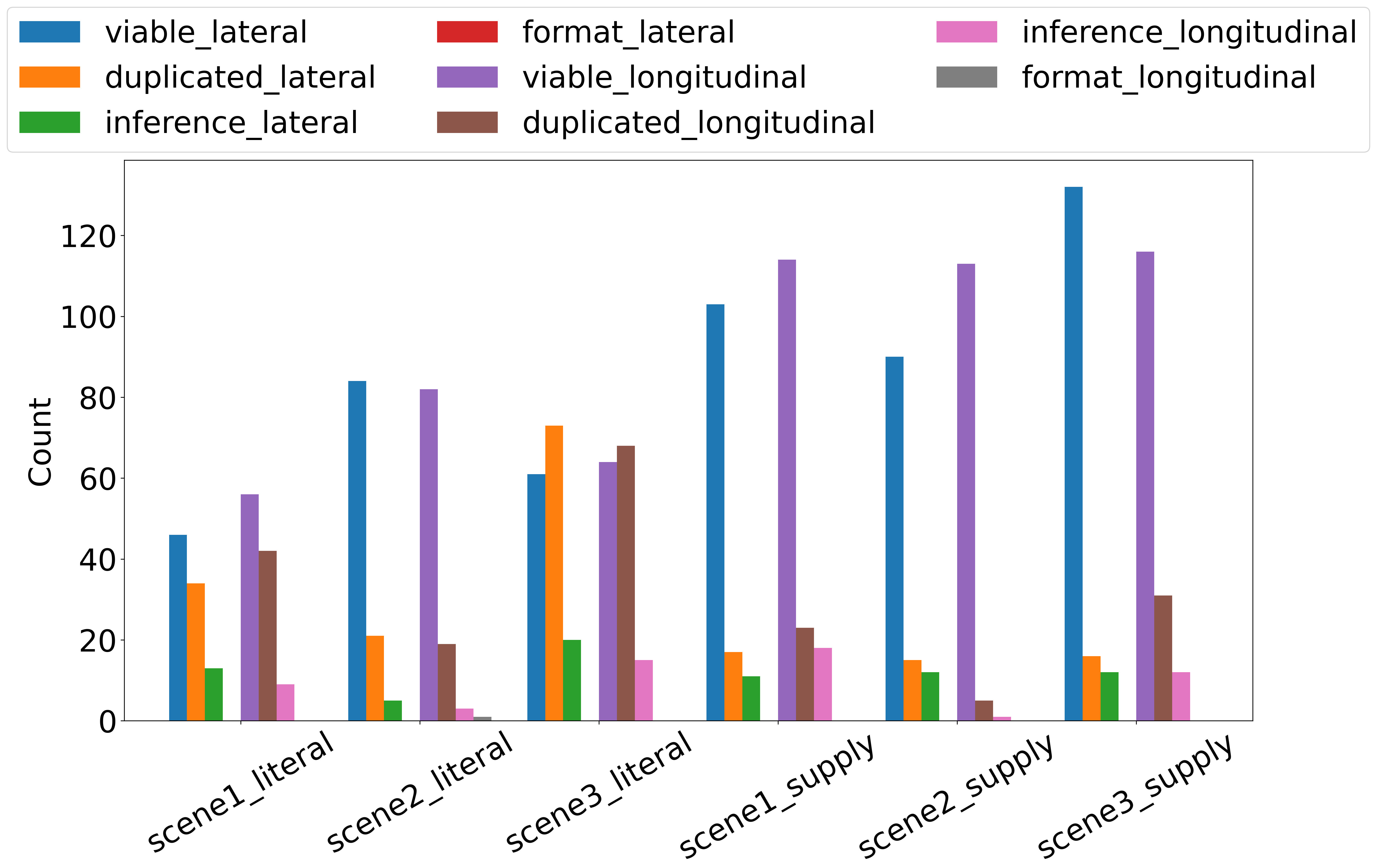}
    \caption{Logic Extraction and Formalization Result.}
    \label{fig:wid2e}
\end{figure}

We adopt the same parameters as in the NL-to-LTL experiment to formalize production rules from NL text. Each interview transcript segment is first input into CriticNL2LTL to get an intermediate LTL representation. The resulted LTL formula is then fed to another LLM and transformed into executable production rule with grounded variables.

Following the standard cognitive driving pipeline, we divide the driving decision-making process into longitudinal and lateral decision-makings. For each intermediate LTL representation, it is formalized into two production rules describing longitudinal and lateral decision-making processes. Moreover, since not every text segment contains complete descriptions of both decision-making processes, we also prompt the LLM to output “$pass$” when it determines that the decision-related information in an LTL formula is incomplete or missing.

After the extraction of logical information and its formalization into production rules, we also analyze the results of our methodology on the interview transcripts across three driving scenarios. To clarify the results, we define the following four types of formalization results and recorded their frequencies across different scenario datasets:
\begin{itemize}
\item \textbf{Format Mismatch}: The formalized production rule cannot be parsed by the inference engine.
\item \textbf{Duplicated Content}: The generated production rule contains the same content as an existing production rule.
\item \textbf{Inference Error}: The generated production rule contains undefined variables or operators.
\item \textbf{Viable}: None of the above three types of errors appear in the generated production rule.
\end{itemize}
As shown in Figure 3, most formalized production rules are viable. The suffixes “supply” and “literal” refer to two different generation methods, which will be clarified later. 

Using the generated production rules, we construct cognitive driving agents for 3 scenarios and train them through the cognitive reinforcement learning pipeline with the collected human driving data. The key of this training is to update the utility of each production in the driving models. To facilitate the training, the inputs of the model are precisely annotated scene elements instead of original sensory data such images and point clouds, and the outputs are the control signals of steering wheel, throttle and brake. A positive reward $R_+$ is given when the model predicts an action consistent with human choice, otherwise a negative reward $R_-$ is assigned. More detailed parameter setting is shown in Table 2.

To distinguish our methodology from current approaches that directly query the LLM how to react under unknown situations, we also collect a set of production rules by prompting the LLM to actively infer and fill in potentially missing environmental information in the interview transcripts. This was done to explore whether further incorporating the LLM’s prior knowledge to discover hidden logic could improve the automated cognitive modeling. Such approach is noted as $M_{sup}$, and the original approach of formalizing production rules literally from the interview transcripts is noted as $M_{lit}$. The cognitive driving models constructed using both methods go through the same cognitive reinforcement learning procedure on the actual human datasets.

\begin{figure*}[t]
\centering
\begin{subfigure}{0.3\linewidth}
  \centering
  \includegraphics[width=\linewidth]{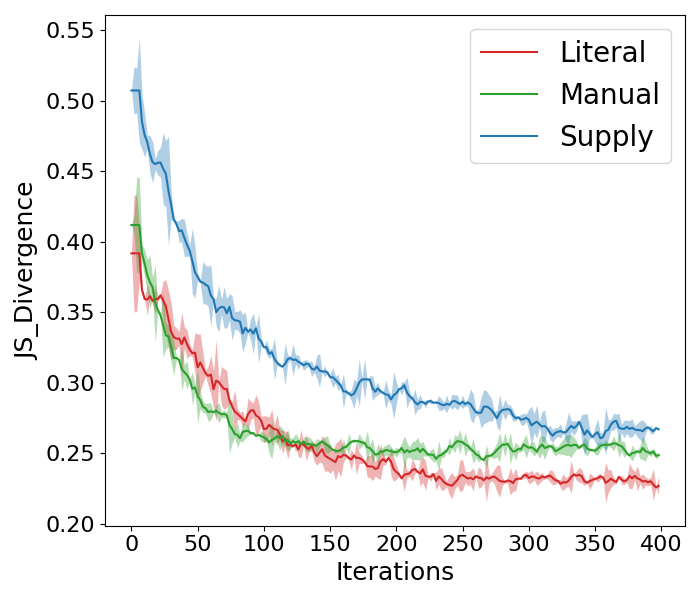}
  \caption{Highway}
  \label{fig:img4}
\end{subfigure}
\hspace{0.01\linewidth}
\begin{subfigure}{0.3\linewidth}
  \centering
  \includegraphics[width=\linewidth]{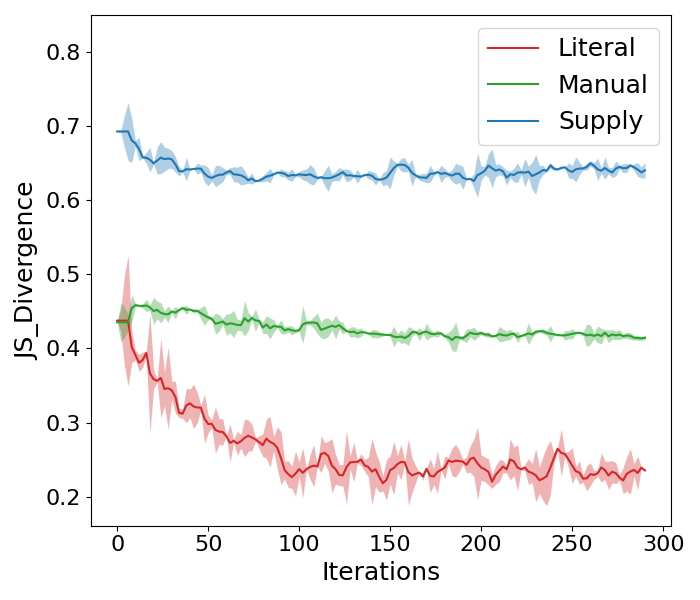}
  \caption{Signalized Intersection}
  \label{fig:img5}
\end{subfigure}
\hspace{0.01\linewidth}
\begin{subfigure}{0.3\linewidth}
  \centering
  \includegraphics[width=\linewidth]{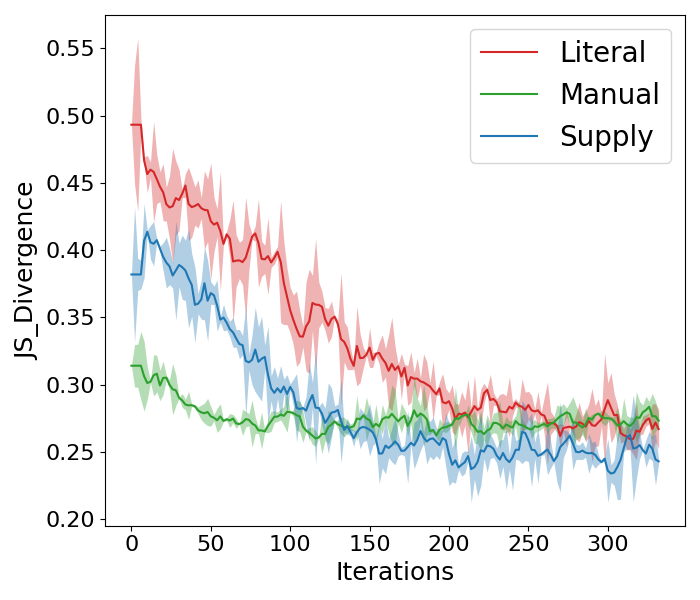}
  \caption{Lane Changing Interference}
  \label{fig:img6}
\end{subfigure}

\caption{JS Divergence Curves of Cognitive Models. }
\label{fig:triple2}
\end{figure*}

After optimizing the cognitive driving models, we evaluate the performance of each model against the test results collected from human subjects in the three driving scenarios. We measure key driving performance metrics including RSR (reasoning success rate), ACR (average completion rate), AST(average simulation time) and ALD (average lane deviation), reported metrics are the average results for 50 test runs.F or more comparative results, test experiments using a manually designed interpretable cognitive driving baseline model \citep{li2024interpretable} are also carried out, shown in Table 3.

As shown in the experimental results, $M_{lit}$ demonstrates competitive performance against the manually constructed baseline that is incorporated with experts' prior knowledge, but $M_{sup}$ shows no advantage against these two methods. The manual baseline, benefiting from experts' dedicated design, exhibits high reasoning completeness and thus holds a distinct advantage in RSR metric. Across all three test scenarios, it achieves an average RSR of $99.47\%$, along with highest average ACR of $93.09\%$. In contrast, $M_{lit}$ achieves an average RSR of $95.69\%$ and an average ACR of $90.53\%$. However, the manual baseline still shows a significant gap compared to $M_{lit}$ in terms of Success Rate(SR) and ALD. Specifically, $M_{lit}$ achieves the highest average SR of $78.57\%$ and an average ALD of $21.54cm$, while the manual baseline scores $65.30\%$ and $24.77cm$, respectively.

To fully evaluate the cognitive models' alignment with human behaviors, we further computed the Jensen-Shannon (JS) Divergence between the decision distributions $P$ generated by each cognitive model and the corresponding distributions $Q$ observed in actual human behaviors. The definition of JS divergence is as follows:
\begin{equation}
JS(P \| Q) = \frac{1}{2} \, KL(P \| M) + \frac{1}{2} \, KL(Q \| M)
\end{equation}
where $KL$ refers to Kullback-Leibler (KL) Divergence and $M = (P+Q)/2$. Specifically, we select the top 10 environmental inputs with the largest number of human decision samples. For each of these inputs, the cognitive decision models are executed $n$ times to generate the corresponding decision distributions. $n$ denotes the number of samples with a specific environmental input. During the training process of the cognitive models, we continuously track the corresponding JS Divergence. The results are shown in Figure 4.

As shown in the results. $M_{lit}$ demonstrates the best alignment with human behavior by achieving the lowest JS Divergence on average, while performing a decent learning capacity. In contrast, the JS Divergence of the manual baseline shows limited change as the number of training iterations increases in Signalized Intersection and Lane Changing Interference scenarios. This indicates that the performance of manual baseline relies more on experts' subjective design, rather than learning from human driving behavior.

$M_{sup}$ demonstrates a certain degree of learning capacity. However, its ability to approximate human decision distributions remains relatively limited. This may be attributed to the fact that the preconditons of its production rules are extended by the LLM, potentially resulting in Over-constraining. Such constraints can lead to a reduced number of decision branches of the cognitive model (fewer decision options for a given environmental input), limiting the model's capacity to fit the human decision distributions.

\section{Discussion}
We propose a method to automatically build cognitive decision rules from natural language texts without human intervention, which can benefit both AI and cognitive computing. For AI, it offers a feasible approach to auto-formalize unstructured texts to formal logic, providing an alternative paradigm different from reasoning all relative texts through a “black box” network, yielding traceable and understandable decision-makings to avoid the hallucination of LLMs caused by their probalistic nature \citep{wang2023drivemlm} \citep{zhang2023survey}. 
For cognitive computing, this method opens a path for low labor-cost automatic modeling based on human natural language, which can help investigating human performances in various human-machine interactive tasks, such as 
cooperative driving or other critical interactive systems.

\section{Conclusion}
This paper presents a framework for automatically formalizing cognitive decision-making from human experience documents and constructing cognitive models for imitation of human behaviors. We demonstrate how this framework could formalize production rules from texts and be applied to driving simulations. This work provides brand new insights on automated cognitive modeling based on NL data.

\section{Acknowledgments}
This work is supported in part by National Natural Science Foundation of China under Grant 62476270 and Grant T2192933.

\bibliography{aaai2026}

\end{document}